\documentclass[letterpaper, 10 pt, conference]{ieeeconf}  

\IEEEoverridecommandlockouts        
\usepackage{graphics} 
\usepackage{epsfig} 
\usepackage{amsmath, amssymb} 
\usepackage{cite}

\usepackage{amsthm}
\usepackage{url}
\usepackage{booktabs} 
\usepackage{hyperref}
\usepackage{multirow} 
\usepackage{subcaption}
\def\x{\mathbf{x}}
\def\a{\mathbf{a}}
\def\s{\mathbf{s}}
\def\o{\mathbf{o}}
\def\h{\mathbf{h}}
\def\g{\mathbf{g}}
\def\p{\mathbf{p}}

\title{\LARGE \bf Learning Visual Feedback Control for Dynamic Cloth Folding}

\author{Julius Hietala, David Blanco-Mulero, Gokhan Alcan, Ville Kyrki
\thanks{This work was financially supported by Academy of Finland grant numbers 328399 and 317020. (Corresponding author: David Blanco-Mulero)}%
\thanks{The authors are with the Intelligent Robotics Group, Department of Electrical Engineering and Automation (EEA), Aalto University, 02150, Espoo, Finland. (e-mail: {\tt\footnotesize julius.hietala@alumni.aalto.fi; david.blancomulero@aalto.fi; gokhan.alcan@aalto.fi; ville.kyrki@aalto.fi}) }%
}

\begin{document}
\newpage
\begin{titlepage}
   \begin{center}
       \vspace*{10cm}

       \textbf{© 2022 IEEE. Personal use of this material is permitted. Permission from IEEE must be
obtained for all other uses, in any current or future media, including
reprinting/republishing this material for advertising or promotional purposes, creating new
collective works, for resale or redistribution to servers or lists, or reuse of any copyrighted
component of this work in other works.}

   \end{center}
\end{titlepage}

\maketitle
\thispagestyle{empty}
\pagestyle{empty}

\begin{abstract}
Robotic manipulation of cloth is a challenging task due to the high dimensionality of the configuration space and the complexity of dynamics affected by various material properties.
The effect of complex dynamics is even more pronounced in dynamic folding, for example, when a square piece of fabric is folded in two by a single manipulator. 
To account for the complexity and uncertainties, feedback of the cloth state using e.g.~vision is typically needed.
However, construction of visual feedback policies for dynamic cloth folding is an open problem.
In this paper, we present a solution that learns policies in simulation using Reinforcement Learning (RL) and transfers the learned policies directly to the real world.
In addition, to learn a single policy that manipulates multiple materials, we randomize the material properties in simulation. We evaluate the contributions of visual feedback and material randomization in real-world experiments.
The experimental results demonstrate that the proposed solution can fold successfully different fabric types using dynamic manipulation in the real world.
Code, data, and videos are available at \url{https://sites.google.com/view/dynamic-cloth-folding}.



\end{abstract}


\section{Introduction}
\label{introduction}

Robotic manipulation of deformable objects has been applied to a different range of tasks from unfolding, folding, and shaping cloths \cite{miller2012geometric, seita_bedmake_2019, ganapathi2020learning, lin2021_vcd} to rearranging objects such as cables, fabrics or bags into a goal configuration \cite{seita_bags_2021}.
A common denominator of these solutions is that they involve static or quasi-static manipulation \cite{Mason583093}. Instead, recent works have explored the effectiveness of dynamic manipulation when applied to deformable objects \cite{ha2021flingbot, jangir2019dynamic}.


Dynamic manipulation can be described as manipulation tasks where inertial forces are a crucial part of the process \cite{Mason583093}.
By contrast, quasi-static manipulation tasks neglect these forces.
There are several benefits of dynamic manipulation over quasi-static manipulation when applied to cloth manipulation.
First, dynamic manipulation enables reaching parts of the configuration space that cannot be reached with quasi-static manipulation.
This can be useful for manipulating parts of the cloth that are not grasped by the manipulator by, for example, performing a dynamic primitive such as flinging the cloth \cite{ha2021flingbot}.
Secondly, dynamic manipulation has been shown to be more efficient in tasks like unfolding cloth, where it requires fewer interactions with the environment than by performing quasi-static manipulation, such as pick and place primitives \cite{ha2021flingbot}.

\begin{figure}
    \centering
    \includegraphics[width=0.99\columnwidth]{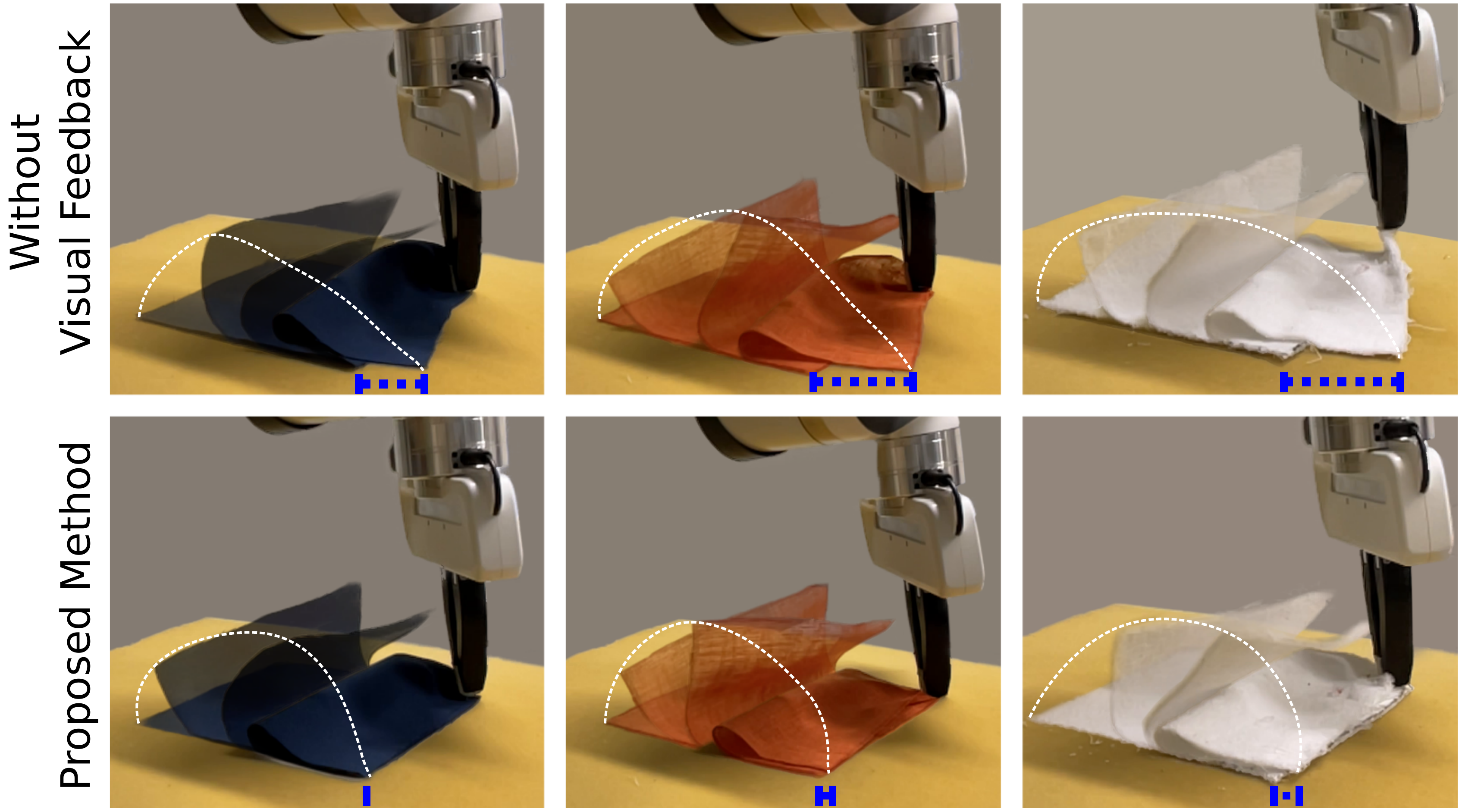}
    \caption{
    The proposed solution succeeds in cloth folding via dynamic manipulation for different types of fabrics when compared to a fixed trajectory from a policy trained in simulation without visual feedback. The dashed blue line indicates the distance from the end position of the cloth non-controlled point to the target point.}
    \label{fig:concept-image}
\end{figure}

In general, dynamic manipulation of cloth in the real world presents the following challenges:  
\begin{itemize}
  \item Observing the position and velocity of the cloth is not straightforward. The methods used in cloth manipulation either observe the cloth visually \cite{lin2021_vcd, matas2018sim} or use a motion capture system with markers attached to the cloth \cite{balaguer2011combining}. In dynamic manipulation, observing the velocity is especially important since the task's success depends on forces caused by acceleration.
  \item The behavior of the cloth depends on the material's physical properties. A fixed trajectory that successfully folds a certain cloth does not necessarily achieve the same end result on a different type of fabric (see Fig.~\ref{fig:concept-image}).
\end{itemize}
Dynamic cloth folding has been studied only in simulation \cite{jangir2019dynamic}, where the true position, velocities, and forces of the cloth can be utilized.
Thus, building visual feedback policies that can manipulate multiple materials in the real world remains an open problem.

In this work, we propose to solve the aforementioned problem by learning end-to-end a visual feedback policy using Reinforcement Learning (RL).
The policy is learned in simulation using Soft-Actor Critic (SAC) \cite{haarnoja2018sac} and then transferred to the real world.
In order to adapt to different fabric types, our work makes use of domain randomization (DR) \cite{tobin_2017_domain_rand, james_2017_transfer_domainrand, petrik_2019_strip_folding}, where we randomize the physical properties of the cloth in simulation, to learn material-agnostic policies.
In addition, we embed a Cartesian controller into the policy learning, where an identical controller is used in simulation and the real world.
This helps the policy to account for the torque commands generated by the controller.
We evaluate the trained policies both in simulation and in real-world experiments.
Our results demonstrate that the visual feedback policies trained with dynamics DR succeed in dynamic cloth folding on a variety of fabrics.
By contrast, visual feedback policies trained without dynamics DR as well as policies executed in an open-loop manner perform poorly across different fabrics.
To the best of our knowledge, this is the first work to learning cloth folding using dynamic manipulation in the real world.

The specific contributions of this paper are then:
\begin{enumerate}
\item A method using domain randomization and visual feedback  to learn policies for dynamic manipulation of different materials,
\item Experimental results showing that visual feedback with domain randomization is crucial for dynamic manipulation,
\item Demonstration of transfer of manipulation policies learned in simulation to the real world without any exploration in the real world. 
\end{enumerate}

\section{Related Work}
\label{related-work}




\subsection{Quasi-static cloth manipulation}
Quasi-static manipulation has been widely studied in cloth manipulation.
Usually quasi-static cloth manipulation consists of finding the suitable pick-and-place position for tasks such as unfolding \cite{lin2021_vcd, seita_bags_2021} or folding \cite{lee2020learning, wu2019learning, weng_2021_fabricflow}.
There is quite an extensive number of methods that have been used to 
predict the pick and the place position ranging from value-based RL \cite{lee2020learning}, model-free RL \cite{wu2019learning} and optical flow \cite{weng_2021_fabricflow}, to learning surrogate models for planning \cite{lin2021_vcd, yan_2020_contrastive_estimation}.
These approaches only consider the start and end state of the cloth, thereby manipulating in an open-loop fashion, where they omit the cloth state while being manipulated.

An alternative is to use visual feedback policies, where the state of the cloth can be estimated at every time step of the manipulation and the policy can react accordingly.
Matas \textit{et al.} \cite{matas2018sim} proposed to transfer from simulation to the real world (sim-to-real) a visual feedback policy to perform quasi-static cloth manipulation tasks.
In our work, we use a similar policy and training architecture. However, we evaluate the learned policies across different fabric materials in a dynamic cloth folding task, where the physical properties of the cloth play an essential role on the manipulation success. In addition, we use SAC instead of Deep Deterministic Policy Gradients (DDPG) \cite{lillicrap_2015_ddpg}.


\subsection{Dynamic cloth manipulation}
Prior works on dynamic cloth manipulation have used dynamic primitives such as a fling motion \cite{ha2021flingbot}, parameterized trajectories \cite{zhang_2021_1d_dynamic, lim2021planar} or visual feedback \cite{yamakawa2011dynamic}.
Ha \textit{et al.} \cite{ha2021flingbot} proposed a fling dynamic motion based on value-based RL to solve cloth unfolding.
However, their approach only takes into account the start and end state of the cloth, thereby not being able to react to changes of the cloth configuration while manipulated. Furthermore, \cite{ha2021flingbot} trains policies in simulation and then collects additional real-world experience to continue the training. 
In our work, we transfer the policies from sim-to-real without any further training using real-world data.
Previous works that have solved dynamic cloth manipulation as an optimization problem \cite{zimmermann2021dynamic}, or using parameterized trajectories \cite{zhang_2021_1d_dynamic}, execute their trajectories in an open-loop manner. Similarly, these solutions cannot adapt their trajectories while the motion is executed or generalize across different cloths.
The use of visual feedback for dynamic cloth manipulation has been previously studied \cite{yamakawa2011dynamic}. However, it has been restricted to custom hardware with no learning involved.
In this work, we propose to learn visual feedback policies that adapt to different fabric types using RL.

Recently, Jangir \textit{et al.} \cite{jangir2019dynamic} studied three dynamic cloth manipulation tasks in simulation, where they proposed to learn a policy using RL.
Their work was limited to simulation, where the policy used the internal states of the cloth given by a simulator to predict the next action. This is infeasible in the real world as the true state of the cloth cannot be measured.
In this work we study one of the dynamic manipulation tasks proposed by \cite{jangir2019dynamic}, that is, sideways folding. In contrast, we tackle the challenge of observing the cloth in the real world by learning a visual feedback policy.

\section{Background}
\label{sec:background}
\subsection{Problem statement}
\label{sec:problem-statement}

We consider a scenario, where a robotic manipulator interacts in finite episodes over discrete time steps with a cloth that needs to be manipulated to a goal configuration. We then address the problem of how to learn the control policy for the system.
We assume that the full state of the cloth cannot be observed directly in the real world. Instead, we can only  observe its state from visual feedback. Therefore, we formulate the dynamic cloth manipulation problem as a Partially Observable Markov Decision Process (POMDP).
The goal of the RL agent is to find the optimal policy $\pi^*$, parametrized by $\theta^*$, that maximizes the expected value of the cumulative reward

\begin{equation}
    \pi^{*} = \operatorname*{argmax}_{\pi} \mathbb{E}\left[\sum^{N}_{n=1} \gamma \; R(\s_t, \a_t, \g)\right],
\end{equation}
where $\s_t \in S$ set of states, $\a_t \in A$ set of actions, $\o_t \in O$ set of observations, $\g \in G$ set of goals and $\gamma \in (0, 1]$ is the discount factor.
We include a goal $\g$ at each episode which is used to determine whether the configuration of the cloth results in success in the task or not.


\begin{figure*}
    \centering
    \vspace{0.2cm}
    \includegraphics[width=2\columnwidth]{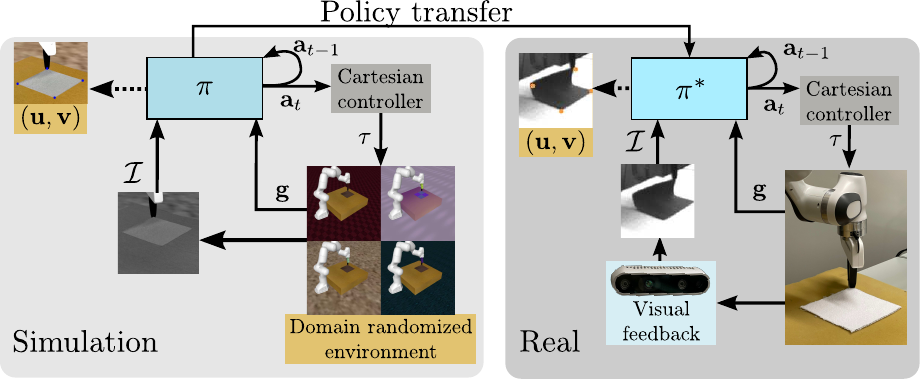}
    \caption{Overview of the proposed solution. The policy is trained in simulation (\textbf{left}), where the simulated environment randomizes the dynamics of the cloth as well as the environment's visual properties. The trained policies $\pi^*$ are transferred directly to the real world (\textbf{right}). The policy receives an image $\mathcal{I}$, a goal $\mathbf{g}$ and the previous action $\mathbf{a}_{t-1}$ as input. The policy outputs the next action $\mathbf{a}_t$ as well as a prediction of the cloth corners $(\mathbf{u}, \mathbf{v})$. Both simulation and real setup use the same Cartesian controller, which generates the torque commands $\tau$ to be executed by the manipulator.}
    \label{fig:struct}
\end{figure*}
\begin{figure}
    \centering
    \includegraphics[width=1.0\columnwidth]{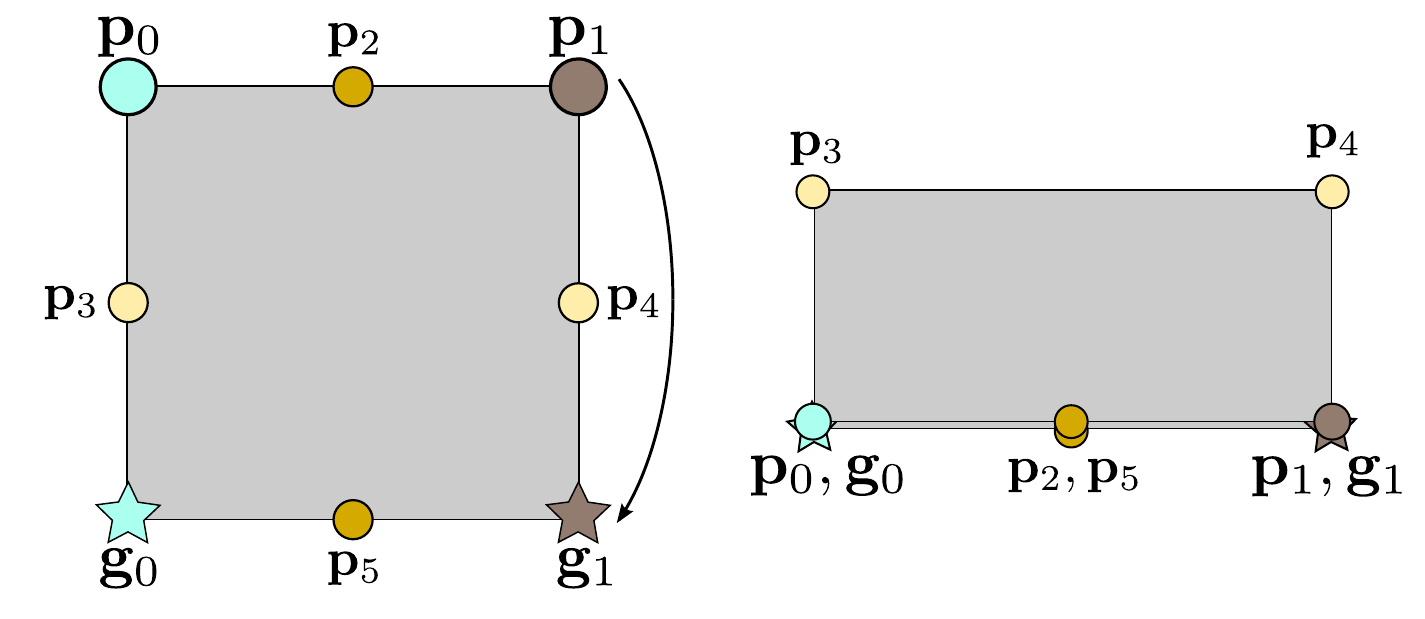}
    \caption{Start (\textbf{left}) and desired (\textbf{right}) end configurations of the cloth for the dynamic sideways folding task, where $\mathbf{p}_1$ is the only controllable point, $\mathbf{p}_0$ is one of the non-controllable corners, and $\mathbf{g}_1, \mathbf{g}_0$ their respective goal locations.}
    \label{fig:start-end-image}
\end{figure}

\subsection{Dynamic sideways cloth folding task}
\label{sec:dynamic-sideways-task-def}
This work focuses on dynamic cloth folding.
The task we consider is dynamic sideways cloth folding, where a single manipulator interacts with the cloth.
This task requires manipulating flat cloths into a folded configuration, where only a single point of the cloth can be controlled.
Thus, the policy needs to exert the necessary forces on the controllable point so that also the non-controllable points get to a folded state.
The starting and folded states are shown on the left and right sides of Fig.~\ref{fig:start-end-image} respectively.
We consider $C=8$ cloth points $(\p_i, \g_i)$, where $\p_1$ is the only grasped corner.
For a fold to be considered successful, the distance
\begin{equation}\label{eq:distance}
    d_\text{sum} = \sum_{i=0}^1 \lVert \p_{i}-\g_{i}\rVert_2 ,
\end{equation}
should be below a threshold $\delta$ at the end of an episode. 

The reward is a combination of a sparse and a dense reward \cite{plappert2018multigoal}. Thus, the agent will get a sparse reward $r_t=-1$ when the fold is unsuccessful, and a dense reward when it is successful according to $r_t = d_\text{sum}$,
where the reward scales linearly from 0 to 1 based on the distances.

In order to focus on learning a visual feedback dynamic manipulation policy we assume that the cloth starts from a grasped state, which can be found using some of the methods discussed earlier \cite{ha2021flingbot, lee2020learning}.

\section{Proposed Method}
\label{sec:proposed-method}
A general view of our proposed method for dynamic cloth folding is depicted in Fig.~\ref{fig:struct}.
The policies are trained in simulation using RL, where we use DR including the physical properties of the cloth, in order to achieve generalization across different fabric types.
In both the sim (left Fig.~\ref{fig:struct}) and real system (right Fig.~\ref{fig:struct}) the policies take a visual input $\mathcal{I}$ and output the next actions to the same Cartesian controller, Operational Space Control (OSC) \cite{khatib1087068}, which facilitates transferring the policy trained in simulation.

\subsection{Learning in Simulation} \label{method:learning-algorithm}

In order to efficiently gather training data for policies that run in the real world, we use a physics engine simulator. The simulation environment includes a flat cloth and a robotic manipulator as shown in Fig.~\ref{fig:struct}. 
The cloth is simulated as a 2D grid of bodies connected via a spring-damper system  \cite{mujoco}, where one of the corners is connected to the robot end-effector to assume the initial grasped state of the cloth.


The simulation provides the internal states $S$ of the cloth, along with a 100$\times$100 grayscale image $\mathcal{I} \in O$ of the cloth.
The images serve as the partial observations of the RL problem and are used by the policy $\pi$ to determine the next action $\mathbf{a}_t$.
In addition, the policy takes as input the previous action $\a_{t-1}$ to provide temporal information, and a goal $\g$ that represents the desired cloth configuration.

We follow the same architecture and training scheme as \cite{matas2018sim}, where the policy is composed of three mappings
\begin{equation}
 \pi:
\begin{cases}
   \h_1: \mathcal{I} \rightarrow \mathcal{L}, \\
   \h_2: (\mathcal{L}, \a_{t-1}, \g) \rightarrow \mathbf{a}_t, \\
   \h_3: \mathcal{L} \rightarrow (\mathbf{u}, \mathbf{v}).
\end{cases}
\end{equation}
First, the input image is mapped to a latent space $\mathcal{L}$ via $\h_1$.
The second mapping $\h_2$ outputs the continuous action $\a_t$, which is the desired change in the end-effector position.
The last mapping $\h_3$ provides an auxiliary output with the estimated position of cloth points $(\mathbf{u}, \mathbf{v})$ in the image plane. The auxiliary outputs are only used during training for guiding the policy to detect relevant features \cite{matas2018sim}.
The first mapping is implemented as a convolutional neural network, whereas the other two mappings are fully connected neural networks.
The inference speed of the policy is crucial for the manipulator to execute the fast dynamic actions.
We describe the network architecture, implementation, and the inference speed in Section~\ref{sec:exp-sim-setup-training}.

We use SAC for learning the visual feedback policy.
Additionally, our method uses Hindsight Experience Replay \cite{andrychowicz2017hindsight} to speed up training.
We also incorporate expert demonstrations in the replay buffer.
The specific details about the policy training are given in Section~\ref{sec:exp-sim-setup-training}.
In our system, the Q-functions of SAC observe the position and velocity of $C$ cloth points $\{ \mathbf{p}_i, \mathbf{\dot{p}}_i\}^C_{i=1} \in S$, along with the goal and current action. The reason behind the observation asymmetry between policy and Q-functions is that the cloth position and velocity information are not available in the real world, and the Q-functions are only needed during training.
The fact that the policy observes only images rather than cloth points facilitates transferring the policy to the real world.

\subsection{Sim-to-real dynamic manipulation} \label{method:sim2real}
In order to effectively transfer a visual feedback policy from simulation we incorporate two methods into the policy learning process:
1) embedding the Cartesian controller used in the real hardware into the policy learning process, and 2) domain randomization, including the physical and visual properties of the cloth as well as a temporal delay in the observations \cite{rupam_2018_rl_task_real}.


The first element of our method that facilitates the transfer is employing the controller of the real manipulator in the policy learning process in simulation.
For performing  dynamic cloth manipulation in the real world we use a cascaded structure combining a low-frequency policy with a Cartesian controller operating at a high frequency required by the real hardware.
The visual feedback policy is executed at a lower frequency due to the real-world constraints at which the images can be observed (see Section~\ref{sec:experimental-setup}).
The Cartesian controller maps the desired positions of the end-effector given by the policy into the desired joint torques of the robotic manipulator.
We propose to use the OSC as Cartesian controller, whereas other types of Cartesian controllers could be used as well.
By embedding the controller in the learning process we let the policy implicitly learn the dynamics of the controller. Therefore, there is no need to tune the gains of the controller to a specific task prior to training.
We use the policy output to compute the target end-effector position $\x_{t+1} = \x_t + \mathbf{a}_{t}$, where $\x_t$ is the previous desired position of the end-effector.
Then, to provide a high-frequency reference signal to the OSC controller in between policy steps, the policy is interpolated $K$ times from $\x_t$ to $\x_{t+1}$ according to

\begin{equation}\label{eq:OSC_interp}
    \x_{d, k+1} = \lambda \cdot \x_{t+1}  + (1-\lambda) \cdot \x_{d,k},
\end{equation}
where $\lambda$ is a filter value controlling the speed of the updates, $\x_{d, k}$ is the current desired position at sub-step $k$, and $\x_{d, 0}$ is the previous end-effector position. 
The reason to use such filtering is to avoid drastic movements that may cause the robot to become unstable when new actions are performed.


The second element to assist in the sim-to-real transfer is DR.
In order to randomize the physical properties of the cloth we first sample the cloth parameters $\xi_c$ from a uniform distribution $\mathbf{\xi}_c \sim \mathcal{U}$ across the possible ranges for each parameter (see Section~\ref{sec:experimental-setup}).
The cloth parameters are simulator-dependent, in our case the spring-damper system is characterized by its impedance, stiffness and damping \cite{mujoco}.
Then, we collect expert dynamic motion demonstrations from the real world that succeed on sideways folding on various real-world cloths (see Section~\ref{sec:exp-sim-setup-training}).
The sampled parameters $\mathbf{\xi}_c$ are evaluated in simulation under the demonstration trajectories.
Out of the sampled parameters, we pick the top $M$ parameter sets that achieve the highest cumulative reward.
Thus, we assume that the parameters that lead to a folded state and the highest cumulative reward are more realistic and can be used for randomizing the physical properties of the cloth.
This approach is opposed to uniformly sampling the physical properties for each episode, which might lead to unrealistic behavior of the cloth that is not useful in the sim-to-real transfer. 
We denote the set of valid cloth properties as $F_{{sim}} = \{f_1, ..., f_M\}$, where we sample from $F_{{sim}}$ at each episode during training.
These properties do not necessarily match the real-world cloth dynamics, as we measure the success of the collected trajectories on a cloth type over different randomized parameters.

In addition to the physical properties of the cloth, we randomize the visual properties of the simulation environment including lighting conditions, image noise, camera position, and object textures. Additionally, we randomize the texture of the cloth, which is sampled from a set of textures gathered from a set of real cloths $F_\text{real}$. These parameters are sampled from a uniform distribution for each episode (see Section~\ref{sec:experimental-setup} for details).

Finally, considering that in the real world the frequency at which we obtain image observations is not stable, we incorporate Gaussian noise in the simulation image sample frequency. Therefore, the policy is trained with a random temporal delay between the taken action and the next observed state. These delays present a high impact on the agent performance \cite{dulac_2021_challenges_realworld_rl}.
Extensions of our work can incorporate a correction of the delay to improve further the agent performance \cite{bouteiller_2021_randomdelays}.




\label{sec:method}

\section{Experiments}
\label{sec:experimental-setup}
%
The goal of the experiments is to evaluate the proposed solution for learning visual feedback policies for cloth folding via dynamic manipulation.
To that end, our experiments investigate:
1) how well does the performance of the policies in simulation match the real world, and
2) the performance of visual feedback policies across different fabric materials.

\subsection{Baselines}\label{sec:exp-setup-baselines}

In simulation, we use as a baseline the method proposed by Jangir \textit{et al.} \cite{jangir2019dynamic}, where the policy directly observes the cloth positions and velocities $(\p_i, \dot{\p}_i)$ and there is no DR in the training process.
We use SAC instead of DDPG so as to compare the policy mapping instead of the policy training method.
The policies trained with this baseline are not directly transferable to the real world since the internal states of the cloth are not available.
Therefore, we train multiple policies, each  using different parameters $\xi_c \sim F_\text{sim}$, and select the ones that achieve the highest cumulative reward.
Then, we execute the fixed trajectories from simulation on the real hardware, which we denote as $\pi_\text{fixed}$ on the real experiments.
This serves also to evaluate the performance of fixed trajectories across cloths with different dynamics.

The second baseline ($\pi_{\text{visual}}$) follows the same training and controller architecture as the proposed solution ($\pi_{\text{ours}}$).
However, the policies are trained without randomizing the cloth's physical properties.
Similarly to the first baseline, we train multiple policies on different parameters and transfer the ones that perform best in simulation.
The policy $\pi_{\text{visual}}$ serves as an ablation study of randomizing the cloth dynamics for learning visual feedback policies. 

\subsection{Simulation setup and training}\label{sec:exp-sim-setup-training}
The simulated environment is implemented in MuJoCo \cite{mujoco} where a model of the Franka Emika Panda is included along with the 2D grid of bodies representing the cloth.

In order to determine the set of cloth physical properties $F_\text{sim}$ we used three different test cloths $F_\text{real} = \{f_{\text{orange}}, f_{\text{white}}, f_{\text{blue}}\}$.
The cloth properties can be qualitatively differentiated as $f_{\text{blue}}$ the most rigid and $f_{\text{orange}}$ as the most shallow and most susceptible to deformation when manipulated (see Table~\ref{tab:fabrics}).
Alternatively, other simulators that support different materials can be used \cite{wang_2011_arcsim}.
The randomization set $F_\text{sim}$ is composed of $M=20$ cloth parameters which are determined using the demonstration-based parameter identification scheme defined in Section~\ref{sec:method}.

\begin{table}
\renewcommand{\arraystretch}{1.3}
\vspace{0.15cm}
\caption{Real-world fabrics set $F_{real}$ used for collecting expert demonstrations and real-world experiments.}
\label{tab:fabrics}
\centering
\begin{tabular}{c  c  c} 
 \toprule
 \bfseries Cloth & \bfseries Fabric type & \bfseries Weight ($\mathbf{g/m^2}$) \\ [0.5ex] 
 \toprule
Blue & Cotton & 400 \\
 \midrule
 White & Cotton & 345\\
 \midrule
Orange & Polyester & 135 \\
 \bottomrule
\end{tabular}
\end{table}

The environment episode has a length of 250 time steps, where the OSC controller acts at the simulation frequency of 100 Hz and the policy at 10 Hz.
This was designed in order to mimic the real system, where the policy is interpolated following \eqref{eq:OSC_interp}.
An episode is terminated either at the end of the time steps or when the cloth corner $\p_1$ is within the threshold $\delta = 4$cm from its goal for more than 100 time steps. 
At each training episode the cloth goal corner $\g_i$ is sampled within 2cm of the adjacent corners to $\p_0$ and $\p_1$.
The goal is specified in Cartesian coordinates as the relative position to the grasped corner given the cloth size.

The policy training is performed for 100 epochs, where each epoch consists of $10^4$ environment time steps.
At the end of each cycle, the collected observations are added to a replay buffer and policy optimization is performed for 1000 gradient steps using SAC. 
The replay buffer is a rolling buffer with $10^5$ buffer size.
We collected a single demonstration and added Gaussian noise $\mathcal{N}(0,0.005)$ to the trajectory. The trajectory is then used to generate a tuple of expert state-actions added to the replay buffer.
About 10\% of the trajectories that are added consist of real-world demonstrations that succeed in cloth folding.
This was done to further aid exploration when training the policy.
Each epoch is concluded with 20 evaluation episodes where the success rate and corner distances to the goals are measured.

The policy network architecture as well as the specific range of randomization parameters can be found in the open source code\footnote{\label{foot:git}https://github.com/hietalajulius/dynamic-cloth-folding}.
The policy output actions are scaled to the range of [-0.03, 0.03] for each dimension.



\subsection{Real-world setup}\label{sec:exp-real-setup}

The real-world experiments are performed using the Franka Emika Panda robot and Robot Operating System\cite{ros}.

The visual feedback input is provided by a Realsense D435 camera.
In order to obtain observations at a valid frequency, we use the maximum available frequency setting of 300 frames per second, where the expected latency is between 25-39 ms \cite{realsense-highspeed}.
The system is designed so that the policy acts at 10 Hz and the latest camera observation is given to the policy.
The chosen frequency ensures that each action is determined based on the cloth behavior that took place after the previous action, which is a requirement for the Markov property to hold \cite{shi2020does}.
This procedure is followed for both $\pi_\text{visual}$ and $\pi_\text{ours}$, whereas the fixed trajectory policy $\pi_\text{fixed}$ sends directly the trajectory via-points at a frequency of 10 Hz to the OSC controller. 
The policy is implemented using PyTorch \cite{paszke_2019_pytorch}, which allows for an inference time below 20 ms.

In order to validate the performance over different cloth types, the real-world experiments are run across the set of $F_\text{real}$, where we run 10 evaluations per each fabric and policy type.
The baseline policies $\pi_\text{fixed}$ and $\pi_\text{visual}$ transferred to the real world are selected as those that perform best under the cloth parameters $\xi_c$ that match better the demonstration trajectories for each of the real test cloths.


\begin{figure*}
    \centering
    \vspace{0.15cm}
    \includegraphics[width=2\columnwidth]{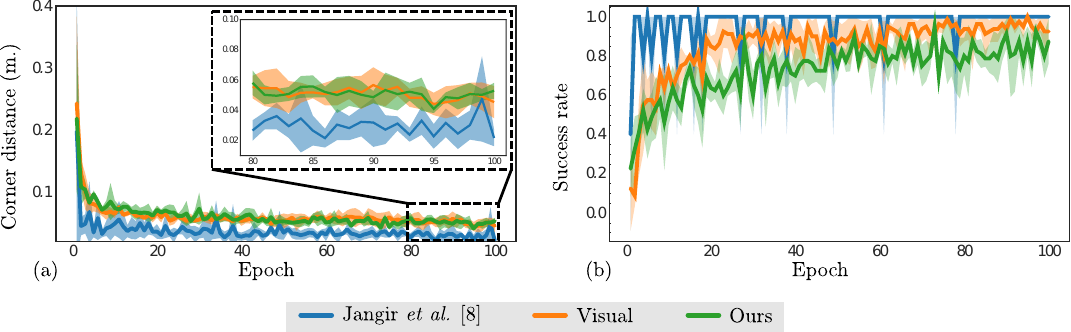}
    \caption{Mean and variance for (a) fold distance error $d_\text{sum}$ and (b) success rate for each evaluation epoch for each method.}
    \label{fig:sim-evaluation}
\end{figure*}


Prior to running an evaluation, the cloth to be folded is attached to the robot's end-effector and the robot is reset to the same joint configuration for each episode.
The grasping, camera configuration and lighting conditions are kept fixed across all evaluation types to make the evaluation as standardized as possible.
The evaluation includes empirically measuring the distance of $\p_0$ and $\p_1$ to their goal locations $\g_0$ and $\g_1$ at the end of the trajectory.
The goals are provided to the policies as in the simulated setup.

\section{Results}
\label{sec:results}

We evaluate the folding performance of the different policies by measuring the sum of the distances \eqref{eq:distance} of the grasped and non-grasped corners to their goals once each trajectory has finished.
It should be noted that $d_\text{sum}$ was used in the reward function during training, so this measure serves also to assess the sim-to-real transfer quality.
To determine whether any difference in the evaluation metrics between the methods is statistically significant we use the Mann-Whitney U test for the real-world experiments (Section~\ref{sec:results-real}).
We choose a statistical significance level of 0.05 prior to the evaluations to assess whether the underlying distributions of the corner distance measurements across evaluations differ at least by the amount reported.





\subsection{Simulation results}
\label{sec:results-sim}
First, to set a baseline for the real-world folding task,
we assess the policy evaluation during training
over five different random seeds for each method via the metric $d_{\text{sum}}$ and the success rate, Fig.~\ref{fig:sim-evaluation}a and Fig.~\ref{fig:sim-evaluation}b respectively.

The results show that the policies can converge and succeed on the task within 100 epochs.
The policy based on the work by Jangir \textit{et al.} \cite{jangir2019dynamic} is able to learn the task only after a few epochs.
We hypothesize that the success rate converges faster than in what is reported in \cite{jangir2019dynamic} due to the fact that we omit the grasping problem.
Additionally, the use of SAC is expected to aid in exploration compared to DDPG used in \cite{jangir2019dynamic}.
The corners error $d_{\text{sum}}$ ranges from 2-4 cm on average at the end of the training, which is approximately $50\%$ less than the visual feedback-based methods.
This result suggests that observing the exact positions and velocities of selected points on the cloth can attain more accurate folds than the visual observation methods.

\subsection{Real-world results}
\label{sec:results-real}

\begin{figure}
    \centering
    \includegraphics[width=\columnwidth]{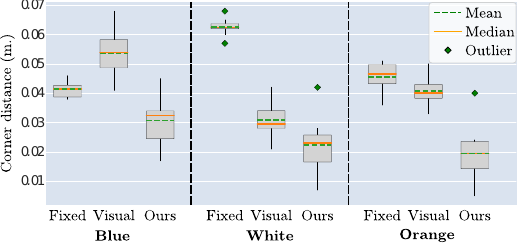}
    \caption{Real-world measurements of distance error $d_\text{sum}$ (in m) for the different methods.}
    \label{fig:real-results}
\end{figure}

\begin{table}
\renewcommand{\arraystretch}{1.3}
    \caption{Statistical significance (p-values) of differences between Ours against the evaluated baseline methods using the Mann-Whitney U test. All results are statistically significant ($p < 0.05$).}
    \label{table:2}
    \centering
    \begin{tabular}{c  c  c }
         & Fixed & Ours-  \\ [0.5ex]
         \toprule
        Ours (Orange) & $\mathbf{4 \cdot 10^{-5}}$ & $\mathbf{2 \cdot 10^{-4}}$ \\
        \midrule
        Ours (White) & $\mathbf{1 \cdot 10^{-5}}$ & $\mathbf{9 \cdot 10^{-3}}$ \\
        \midrule
        Ours (Blue) & $\mathbf{2 \cdot 10^{-3}}$ & $\mathbf{4 \cdot 10^{-5}}$ \\
         \bottomrule
    \end{tabular}
\end{table}

First, in order to compare against the simulation results, we evaluate the real-world error $d_\text{sum}$ for each fabric and policy (see Fig.~\ref{fig:real-results}).
The first thing to note is that the proposed solution $\pi_{\text{ours}}$ significantly outperforms the other methods consistently for all fabrics. Furthermore, the achieved error $d_{\text{sum}}$ even falls below the range seen during evaluation in simulation.
The fixed trajectories from $\pi_{\text{fixed}}$ present a significantly higher error compared to simulation for all the cloth types, which highlights the sim-to-real gap.
This can also be seen from the variability across the different cloth types.
The visual feedback-based method $\pi_\text{visual}$ performs considerably better compared to trajectories from $\pi_{\text{fixed}}$ for the white cloth, but is comparable or worse for the other types.
This indicates that visual feedback is not enough for solving the task across a variety of fabrics, and the randomization of the cloth's physical properties plays an essential role while training the policy.
The variance of the feedback-based methods shows that the results are less stable.
This points to the fact that the policies are susceptible to noise from the environment such as lighting and camera position. Comparing the proposed method error to the baselines shows statistical significance for all cloth types, as shown in Table~\ref{table:2}.

\begin{figure*}
    \vspace{0.1cm}
     \centering
     \begin{subfigure}[b]{0.28\textwidth}
         \centering
         \includegraphics[width=\textwidth]{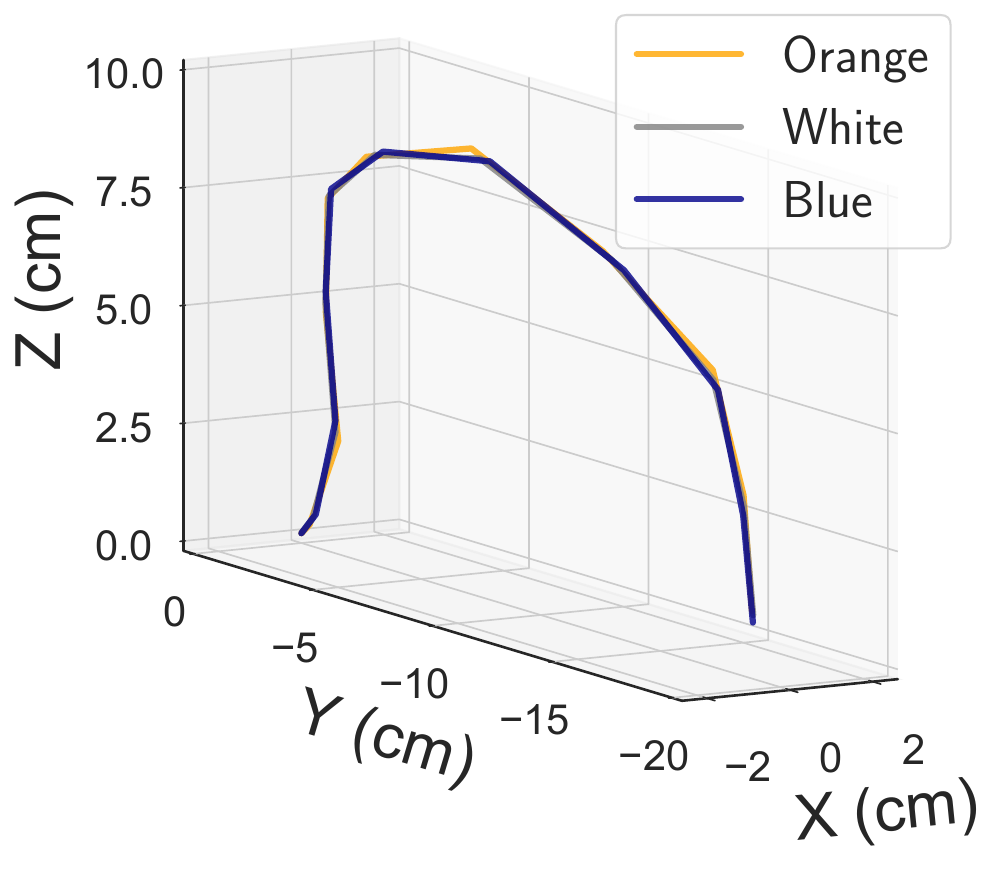}
         \caption{Fixed}
     \end{subfigure}
     \hfill
     \begin{subfigure}[b]{0.28\textwidth}
         \centering
         \includegraphics[width=\textwidth]{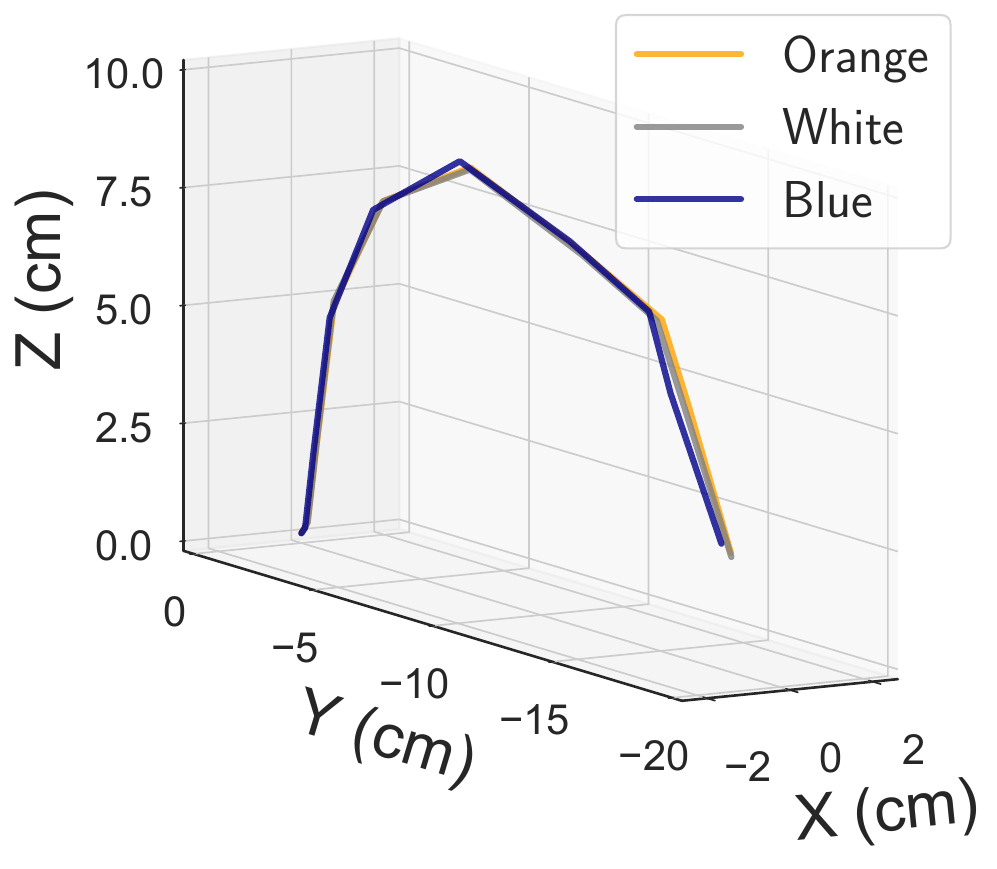}
         \caption{Visual}
     \end{subfigure}
     \hfill
     \begin{subfigure}[b]{0.28\textwidth}
         \centering
         \includegraphics[width=\textwidth]{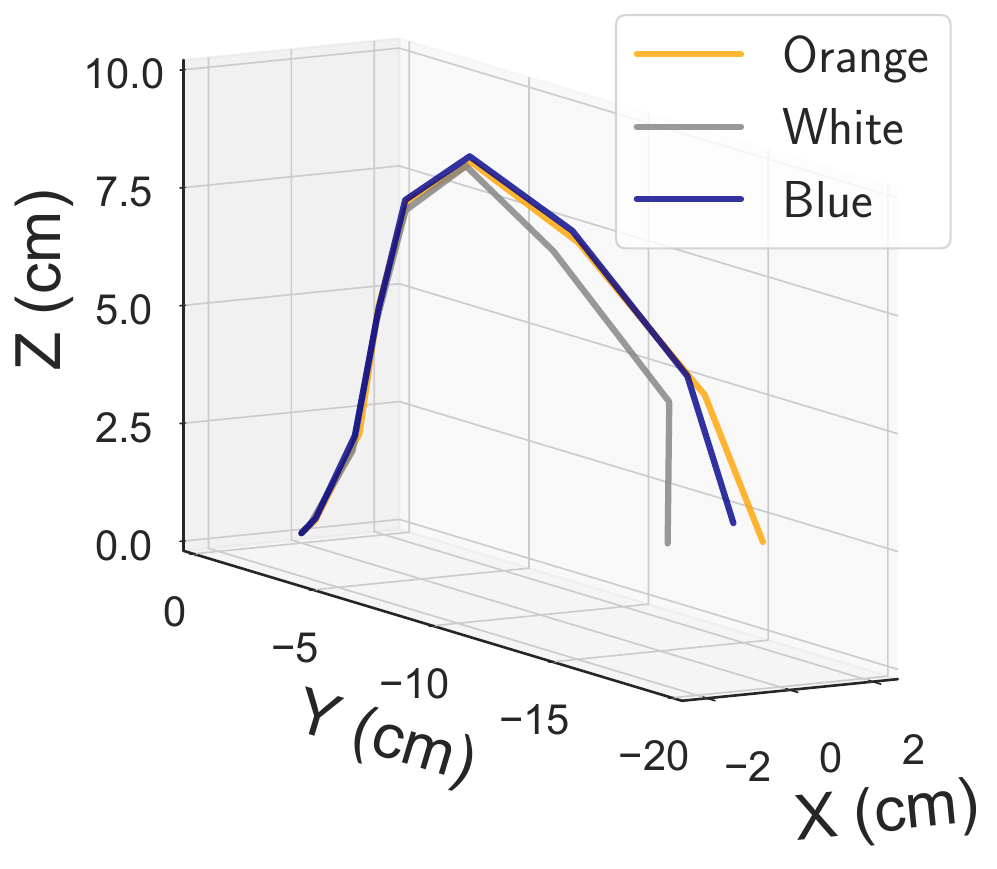}
         \caption{Ours}
     \end{subfigure}
        \caption{Qualitative comparison of mean of dynamic manipulation trajectories from real world  experiments on three fabric types for the policy (a) Fixed, (b) Visual and (c) Ours.}
        \label{fig:real-trajectories}
\end{figure*}

\begin{figure}
    \centering
    \subfloat[Fixed]{\includegraphics[width=\columnwidth]{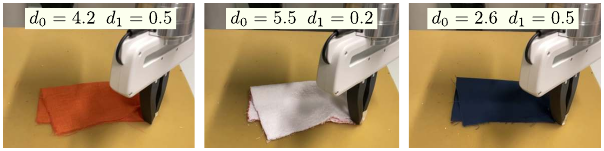}}\\
    \vspace{0.1in}
    \subfloat[Visual]{\includegraphics[width=\columnwidth]{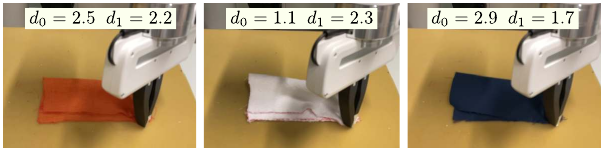}}\\
    \vspace{0.1in}
    \subfloat[Ours]{\includegraphics[width=\columnwidth]{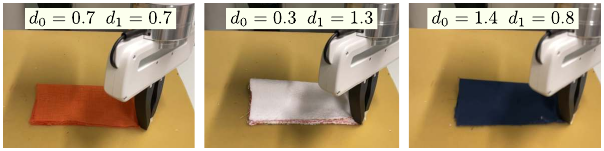}}
    \caption{Qualitative and quantitative results showing the end configurations for each cloth type after a fold for each policy, where $d_0$ and $d_1$ represent the distance to the left and right corners respectively in cm.}
    \label{fig:empirical-resuts}
\end{figure}

We also evaluate how the policies adapt for each of the fabrics by qualitatively comparing the trajectories for each method and fabric in Fig.~\ref{fig:real-trajectories}.
The trajectories for $\pi_{\text{fixed}}$ are nearly identical, which indicates that the best trajectories for each of the cloth sets $F_\text{sim}$ were quite similar.
The trajectories from $\pi_\text{visual}$ present a minor deviation for different cloth types.
In contrast, the trajectories from $\pi_{\text{ours}}$ showcase more variation over the different fabric types.
This indicates that by randomizing the cloth's physical properties the policy can adapt its trajectory to different fabric types.


Finally, we qualitatively and quantitatively compare the policies fold error for each corner, $d_0$ and $d_1$, over the test fabrics in Fig.~\ref{fig:empirical-resuts}.
The results show that $\pi_{\text{ours}}$ can consistently fold both the controllable and non-controllable cloth points across a variety of materials.
The results from $\pi_{\text{visual}}$ are consistent with the previous results, the fold error is higher when the physical properties of the cloth are not randomized.
The trajectories from $\pi_{\text{fixed}}$ achieve folds where the non-grasped corner reaches past its goal, which can be explained by the lack of feedback.

In summary, our results show the importance of visual feedback for dynamic cloth folding.
Additionally, we provide evidence that visual feedback policies trained in simulation with domain randomization can react and adapt its trajectory when manipulating different fabric materials.
While performing the real-world experiments we noticed that the policies trained using the proposed solution were quite sensitive to the lighting conditions and camera position, despite the domain randomization.
Potential extensions of this work can study the effectiveness of the proposed solution to other tasks such as cloth unfolding.




\label{exp-results}

\section{Conclusion}
\label{sec:conclusion}
We presented a solution for cloth folding using dynamic manipulation by transferring visual feedback policies from simulation to the real world.
The policies are trained in simulation using Reinforcement Learning and transferred directly to the real world.
The transferred policies are capable of folding different fabric types by randomizing the physical properties of the cloth in the policy training.
In addition, the solution embeds a Cartesian controller into the learning process, which facilitates the transfer process.
We evaluated the folding performance in the real world and contrasted the results with the simulation, indicating the sim-to-real gap.
Experimental results showed that visual feedback without domain randomization is not sufficient to manipulate multiple fabric types, while training under different cloth dynamics allows to succeed on the dynamic cloth folding task.


In the future, there will be a need for more general solutions that can be employed in a variety of tasks from bed-making to folding clothes. Thus, we need to study how to enable robotic manipulation across a large range of cloths in terms of sizes, shapes, and dynamics. We foresee that dynamic manipulation will play an important role towards solving these manipulation problems.


\bibliography{references}
\bibliographystyle{IEEEtran}

\end{document}